\pdfoutput=1

\documentclass[11pt]{article}

\usepackage{emnlp2021}

\usepackage{times}
\usepackage{latexsym}

\usepackage[T1]{fontenc}

\usepackage[utf8]{inputenc}

\usepackage{microtype}

\usepackage{graphicx}
\usepackage[inline]{enumitem}
\usepackage{multirow}
\usepackage{booktabs}

\setitemize{noitemsep,topsep=0pt,parsep=0pt,partopsep=0pt}

\usepackage{flushend}

\title{Statistically Significant Detection of Semantic Shifts using\\Contextual Word Embeddings}

\author{Yang Liu \\
  University of Helsinki\\
  \texttt{yang.liu@helsinki.fi} \\
  \And
  Alan Medlar \\
  University of Helsinki\\
  \texttt{alan.j.medlar@helsinki.fi}\\
  \AND
  Dorota G\l owacka \\
  University of Helsinki\\
  \texttt{dorota.glowacka@helsinki.fi}
  }

\begin{document}
\maketitle

\begin{abstract}

Detecting lexical semantic change in smaller data sets, e.g.~in historical linguistics and digital humanities, is challenging due to a lack of statistical power. This issue is exacerbated by non-contextual embedding models that produce one embedding per word and, therefore, mask the variability present in the data. In this article, we propose an approach to estimate semantic shift by combining contextual word embeddings with permutation-based statistical tests. 
We use the false discovery rate procedure to address the large number of hypothesis tests being conducted simultaneously.
We demonstrate the performance of this approach in simulation where it achieves consistently high precision by suppressing false positives. We additionally analyze real-world data from SemEval-2020 Task 1 and the Liverpool FC subreddit corpus. We show that by taking sample variation into account, we can improve the robustness of individual semantic shift estimates without degrading overall performance.

\end{abstract}

\section{Introduction}


Semantic change detection methods are used in historical linguistics and digital humanities to study the evolution of word meaning over time and in different domains \citep{kutuzov2018diachronic}. While semantic shift estimates have been shown to correlate with simulated \citep{schlechtweg2019wind,shoemark2019room} and manual \citep{schlechtweg2020semeval} annotations of meaning change, to our knowledge, no existing methods attempt to characterize the uncertainty of the estimated semantic shift for each individual word. 
This is especially problematic because semantic change detection is usually based on word embeddings \citep{kutuzov2018diachronic} and recently it has been observed that their stability can vary widely across term frequencies \citep{wendlandt2018factors,antoniak2018evaluating}, implying that many semantic shift estimates are erroneously inflated or underestimated.
Prior studies have addressed this issue by filtering out words that fall below a term frequency threshold. While this approach can remove a majority of false positives, it risks the introduction of false negatives. 
A more robust approach would take into account the sample variation to determine whether there is evidence that an estimated semantic shift is sufficiently different from zero to be considered statistically significant. Unfortunately, non-contextual word embeddings lose this information, flattening all instances of the same term into a single word embedding.

In this paper, we focus on the problem of robust semantic change detection in scenarios where there is either limited data or low frequency terms of interest (e.g.~\citet{del2018short}). 
Our approach is based on contextual word embeddings, such as those produced by BERT \citep{devlin2019bert}, and permutation-based statistical tests. 
%
%
Contextual word embeddings have several advantages over non-contextual embeddings for inferring semantic shift when there is limited data.
First, we can leverage pre-trained models that were trained on large-scale data, encoding prior knowledge of the language.
Second, as contextual word embeddings are generated for every instance of a given word, there is an opportunity to characterize the strength of evidence for each semantic shift using statistical testing. 

The approach we propose in this paper focuses on the application and evaluation of statistical testing in semantic change detection. Our contributions are as follows:

\begin{itemize}
    
    \item We show how to apply statistical significance tests  
    to any semantic change detection method based on contextual word embeddings. 
    To our knowledge, this is the first paper to use statistical testing in the context of individual words in
    semantic change detection.
    
    \item We show in simulation that using permutation tests while controlling the false discovery rate 
    improves precision and scales to estimating the uncertainty for all words in a vocabulary. 
    
    \item We evaluate the impact of statistical testing on overall performance using manually annotated 
    data sets in multiple languages. In a majority of cases, our approach improves performance, 
    resulting in higher Spearman correlations between estimated semantic shifts and annotations.

\end{itemize}

\section{Related Work}

Computational methods for semantic change detection are used 
to compare corpora spanning decades or even centuries. They have been used, for example, to 
analyze historical word usage \cite{hamilton2016cultural} and 
to identify statistical laws of language change \citep{hamilton2016diachronic}.
More recently, there has been increased interest in detecting short-term meaning change, such as novel slang terms,
in Amazon reviews \citep{kulkarni2015statistically}, 
Twitter data \citep{shoemark2019room} and 
specialist online communities \citep{del2018short}.

Prior to the wide-spread use of word embeddings, numerous methods were developed to detect semantic change, including dynamic topic models \citep{blei2006dynamic}, word co-occurrence statistics \citep{gulordava2011distributional} and graph-based methods \citep{mitra2014s}. 
Methods for semantic change detection based on word embeddings exploit their distributional properties
to identify words whose relative position in the embedding space has changed over time, implying a concordant change in meaning \citep{kutuzov2018diachronic}. 
The earliest work in this area was based on 
continuous training, initializing each embedding model with  embeddings from the previous time step \citep{kim2014temporal}. 
Subsequent methods improved performance by 
training independent embedding models for each corpus \citep{kulkarni2015statistically, hamilton2016diachronic}. Embeddings are invariant under rotation and therefore need to be aligned by solving the orthogonal Procrustes problem \citep{hamilton2016diachronic}.
This alignment step can be avoided altogether by, for example, comparing word neighborhoods \citep{hamilton2016cultural} 
or using temporal referencing \citep{dubossarsky2019time}. 


Semantic shift detection with contextual word embeddings is becoming increasingly popular.
\citet{hu2019diachronic} used BERT embeddings to define exemplar representations for pre-defined word senses to track usage over time.
Several methods have side-stepped the need for known word senses by clustering BERT embeddings 
\citep{giulianelli2020analysing, martinc2020discovery}
and shown that clustering-based approaches can scale to the whole vocabulary \citep{montariol2021scalable}.
Another benefit of using contextual word embeddings is that pre-trained models are widely available for many different languages. These models can be used for fine-tuning to perform semantic change detection using more limited data \citep{martinc2019leveraging}.
Lastly, researchers have started to experiment with ensembling multiple types of word embeddings and distance metrics to improve overall performance \citep{kutuzov2020uio, martinc2020discovery}. 

Recently, there have been several benchmarking studies of semantic change detection methods using simulated data \citep{shoemark2019room, schlechtweg2019wind} and manually annotated data sets \citep{schlechtweg2020semeval}. These studies found that variations on the method proposed by \citet{hamilton2016diachronic} performed best, however, methods based on contextual word embeddings were either absent or
the study was based on data where contextual information was partially lost due to shuffling the order of sentences in the corpus \citep{schlechtweg2020semeval}.

In this paper, we use contextual word embeddings to identify statistically significant semantic shifts. 
Statistical significance has not been the subject of much investigation in the semantic shift literature. Indeed, the only approach we are aware of was proposed by \citet{kulkarni2015statistically}, which used bootstrapping to perform change-point detection in time series. However, their method is not applicable to scenarios where there is only two time points 
nor does it take into account the sample variance within each corpus. 
Our approach addresses both of these issues and can be applied to any semantic change detection method based on contextual word embeddings.

\section{Data}
\label{sec:data}

Table~\ref{tab:data} lists the data sets used in this article. 
%
The Liverpool FC corpus collects data from the Liverpool Football Club subreddit from the Reddit online discussion forum. 
The corpus is in English and split into two time periods from 2011-2013 and 2017 \citep{del2018short}.
SemEval-2020 Task 1 
was created to benchmark semantic change detection methods
using two subtasks: binary classification of whether a word sense has been gained or lost (subtask 1) and 
ranking words according to their degree of semantic change (subtask 2).
The data set contains English, German, Latin and Swedish corpora, all of which contain data from two time periods. All sentences were shuffled and the words lemmatized. 
We only used the manual annotations from subtask 2 \citep{schlechtweg2020semeval}. 

\begin{table}[t]
\small
\centering
\renewcommand*{\arraystretch}{1.2}
\begin{tabular}{lccc}
\hline
                & \begin{tabular}[c]{@{}c@{}}Target\\ words\end{tabular} & \multicolumn{2}{c}{\begin{tabular}[c]{@{}c@{}}Corpus size\\ (million tokens)\end{tabular}} \\ \cline{3-4} 
                &                                                        & C1                                          & C2                                           \\ \hline
Liverpool FC    & 97                                                     & 8.5                                         & 11.9                                         \\
SemEval-2020 English & 37                                                     & 6.5                                         & 6.7                                          \\
SemEval-2020 German  & 48                                                     & 70.2                                        & 72.3                                         \\
SemEval-2020 Latin   & 40                                                     & 1.7                                         & 9.4                                          \\
SemEval-2020 Swedish & 31                                                     & 71.0                                        & 110.0                                        \\ \hline
\end{tabular}%
\caption{Number of annotated words and corpus sizes for Liverpool FC and SemEval-2020 Task 1 corpora.}
\label{tab:data}
\end{table}

\section{Methodology}

Our approach uses contextual word embeddings and permutation-based statistical testing to detect semantic shifts in scenarios where data is limited. 
While our approach can be applied to any method based on contextual embeddings, we used the method proposed by \citet{martinc2019leveraging} because of its conceptual simplicity and faster run-time compared to other methods. 
We generated all contextual word embeddings with BERT \citep{devlin2019bert} using the implementation from HuggingFace's Transformers library \citep{wolf2019huggingface}. In all experiments, we used a base version of BERT (the exact pre-trained models used are specified in later sections) with 12 attention layers and a hidden layer size of 768. All parameters were set to the default values used in the Transformers library ver.~2.5.0, unless stated otherwise.

\subsection{Estimating Semantic Shifts}
\label{sec:diff}

Following the approach outlined by \citet{martinc2019leveraging}, we fine-tune a pre-trained BERT model for each data set, combining both time periods. 
%
After fine-tuning, we use the following procedure to generate word representations: 
we feed sentences of up to 512 tokens into BERT and extract contextual embeddings for each token in the sequence. 
Following \citet{devlin2019bert}, we extract embeddings by summing the last 4 encoder layers in the model. 
As BERT uses byte-pair input encoding, not all tokens correspond to individual words \citep{kudo2018sentencepiece}. 
We therefore create contextual embeddings for each word by averaging the embeddings of its constituent tokens.

For each word, we create a non-contextual embedding for each time period by averaging over all contexts for that word.
We estimate semantic shift using the cosine distance between these two non-contextual embeddings.
To assess the uncertainty in these estimates, we calculate {\em p}-values using permutation tests, which are adjusted for multiple comparisons using a false discovery rate procedure (described below). 
%
If the {\em p}-value is greater than 0.05, there is insufficient evidence to reject the null hypothesis of no difference in meaning between time periods and, therefore, any observed differences can be attributed to random sample variation.
%

\subsection{Permutation Tests}
\label{sec:permutations}

Permutation tests are non-parametric significance tests. Non-parametric tests make no assumptions with respect to the underlying sampling distribution of the data.
The goal of the permutation test is to determine whether the observed test statistic (i.e.~the cosine distance) is significantly different from zero (the null hypothesis being that there is no semantic shift between the two time periods). 
Permutation tests generate the sampling distribution by reassigning group labels (i.e.~time periods) to all observations by sampling without replacement. We then recalculate the test statistic between the two randomized groups. 
This procedure is repeated many times, either by enumerating all possible combinations of group assignments or by randomly sampling $n$ permutations. We calculate the {\em p}-value as the proportion of the sampling distribution that is greater than or equal to the observed test statistic and reject the null hypothesis for all {\em p}-values $< \alpha$, where $\alpha$, the significance threshold, is usually set to 0.05. Clearly, this procedure limits the smallest non-zero {\em p}-value to $\frac{1}{n}$. When we cannot calculate the {\em p}-value exactly (i.e.~if all combinations cannot be enumerated), we first use $n=10^3$. If the {\em p}-value is $< 0.05$, we increase $n$ to $10^4$. Finally, if the {\em p}-value is $< 0.005$, we increase $n$ to $10^5$. {\em P}-values with a value of 0.0 are reported as $\frac{1}{n}$.

\begin{figure}
    \centering
    \includegraphics[width=\columnwidth]{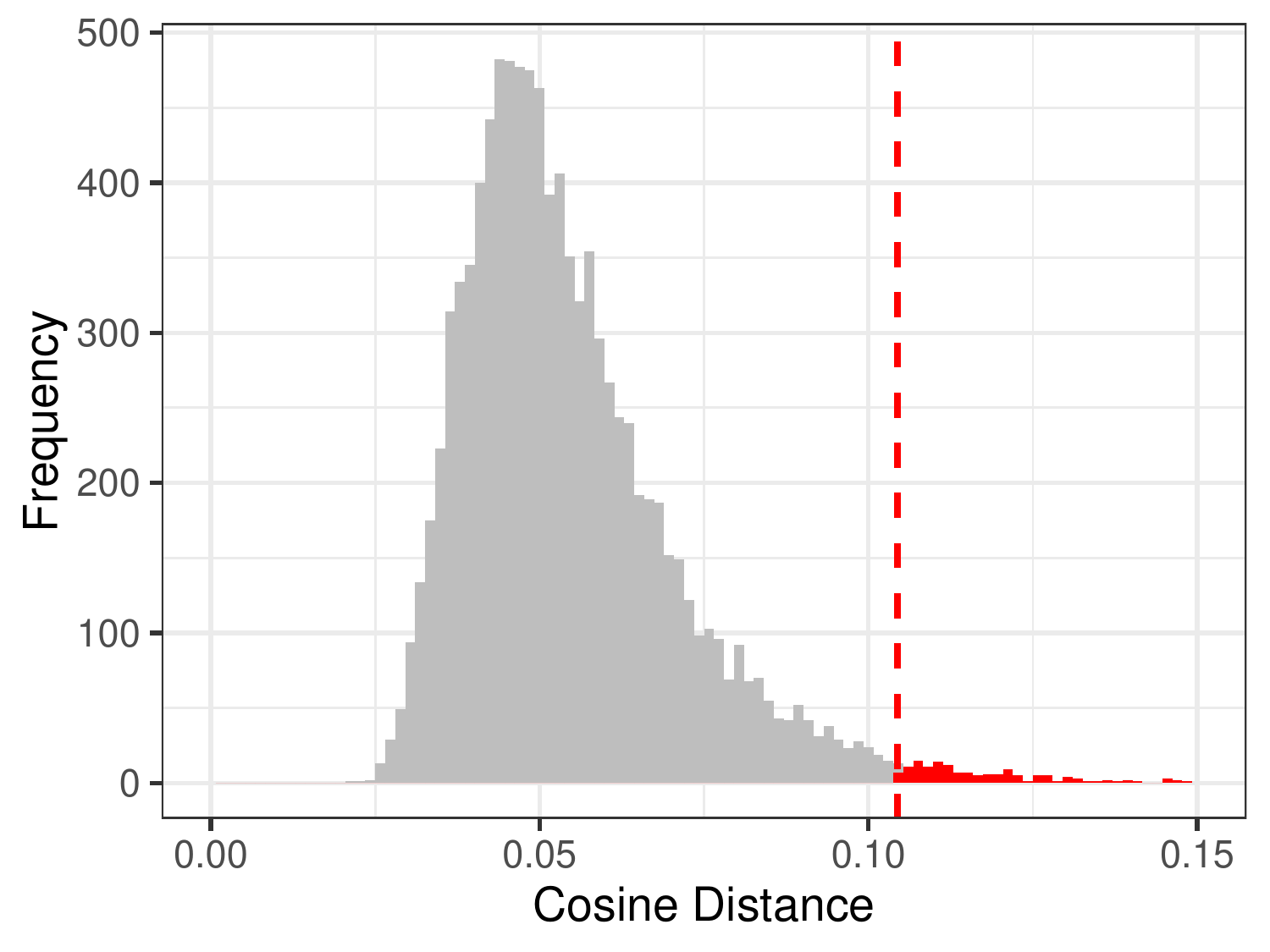}
    \caption{Histogram of the sampling distribution for {\em shovel} from the Liverpool FC data set. The red dashed line at $x=0.104$ is the observed cosine distance between average BERT embeddings. The {\em p}-value is 0.0212 (the proportion of the histogram colored red).}
    \label{fig:perm_example}
\end{figure}

For example, from the Liverpool FC data set, the word {\em shovel} appears 5 and 35 times in the 2011-2013 and 2017 corpora, respectively. The number of combinations, ${40 \choose 5}$, is too large to enumerate exhaustively, forcing us to use random sampling. Figure~\ref{fig:perm_example} shows the sampling distribution together with the observed distance of 0.104 (dashed red line). As 2.12\% of the sampling distribution is greater than or equal to 0.104, the {\em p}-value is 0.0212 and, therefore, statistically significant. 

If we want to investigate many words in the same experiment, however, then we need to consider the issue of multiple comparisons \citep{hsu1996multiple}. In brief, the multiple comparisons problem is where we reject the null hypothesis too often due to the number of simultaneous independent hypothesis tests performed. 
It is, therefore, desirable to set a lower (more stringent) significance threshold for the set of simultaneous significance tests. We do this in a principled way using a false discovery rate procedure.

\subsection{False Discovery Rate}

We use the Benjamini-Hochberg procedure to adjust {\em p}-values for multiple comparisons \citep{benjamini1995controlling}. In doing so, we are limiting the false discovery rate (FDR), i.e.~the proportion of false positives, which can potentially be very large if we perform a significance test for every word in the vocabulary. 

The Benjamini-Hochberg procedure assumes that we perform $m$ significance tests producing a list of {\em p}-values, $P_{1}, P_{2} \dots P_{m}$, ranked into ascending order. We control the FDR at the significance threshold, $\alpha$, by finding the largest value of $k$, such that $P_{(k)} \leq \frac{k}{m} \alpha$. We reject the null hypothesis for all {\em p}-values less than $P_{k}$ \citep{benjamini1995controlling}. 
The implementation of FDR that we use additionally makes the corresponding corrections to all other {\em p}-values, allowing for any $\alpha$ to be used post-correction.

Returning to our previous example, the significance test for {\em shovel} was only one of 97 tests performed to analyze the Liverpool FC data set. Controlling for FDR adjusts the original (significant) {\em p}-value of 0.0212 to 0.0605, suggesting there is insufficient evidence, given the number of independent significance tests performed, to believe the observed distance of 0.104 is different from zero.
We, therefore, attribute the observed difference between time periods to random variation.

\section{Synthetic Evaluation}
\label{sec:syneval}

We created multiple simulated data sets 
to highlight the importance of 
significance testing when performing semantic change detection across the whole vocabulary.
Previous work also used simulated data to compare methods \citep{kulkarni2015statistically, shoemark2019room}, however, they focused on simulating time series, whereas we only simulate two time periods. 

\subsection{Synthetic Data Set Construction}

We create synthetic data sets using a method similar to \citet{shoemark2019room}. For each simulation run, we create two synthetic corpora: $C1$ and $C2$. Each synthetic corpus is created by randomly sampling with replacement 70\% of the sentences from the Liverpool FC 2017 corpus (the larger of the two time periods). $C1$ and $C2$ have the same distributional characteristics, but vary due to sampling noise. We insert controlled shifts by copying and editing sentences in the data set, altering both the term frequency and co-occurrence distributions of shifted words (described below). Finally, we fine-tune the English {\tt BERT-base-uncased} model for 5 epochs using all unique sentences from $C1$ and $C2$, before calculating cosine distances, permutation tests and FDR correction.

Our procedure for inserting semantic shifts is as follows. We select $n$ pairs of words: an acceptor word, that gains new meanings, and a donor word, where these new meanings come from. We pick word pairs by 
\begin{enumerate*}[label=(\roman*)]
\item filtering out words with low ($< 5$) or high ($> 500$) term frequencies and those with $>5$ word senses according to WordNet (very few words in WordNet have only a single word sense, the threshold was chosen to limit the number of word senses without being too restrictive), 
\item we then sort words into descending order by term frequency, 
\item we pair up consecutive words and, 
\item randomly select word pairs. 
\end{enumerate*}
For each word pair, we randomly generate a proportion, $p$, and sample $p$ of the sentences containing the donor word from $C2$ and replaced the donor word with the acceptor word. 
We only simulated gains in word meaning because they are equivalent to losses with only two time points, i.e.~a gain from $C1 \rightarrow C2$ is equivalent to a loss from $C2 \rightarrow C1$. 
To ensure that we do not create any unintentional changes in the meaning of the donor words, these changes were made to copies of the sampled sentences, i.e.~the term frequency of donor words is unchanged.

We created 10 data sets using this procedure, each simulating 500 semantic shifts. Given that $p$ is randomly generated, these shifts can result in anything from a term frequency gain of 1 (which is undetectable due to sampling variation) to the term frequency approximately doubling. 
As a result of the initial random sampling to create the corpora,
the number of non-artificially shifted words with a difference in term frequency of at least +50\% was $\mathord{\sim}1200$ and +100\% was $\mathord{\sim}$150.
The size of the shared vocabulary between $C1$ and $C2$ was $\mathord{\sim}$31,000 words in all simulation runs. 

Model fine-tuning took $< 2$ hours on an Nvidia Volta V100 GPU with 32GB of RAM. {\em P}-value calculations were parallelized, taking a total of 170 CPU hours  per simulation run on Intel Xeon CPUs running at 2.1 GHz.

\subsection{Baseline Optimization}
 
As a baseline, we compared our approach to the method proposed by \citet{martinc2019leveraging}, described in Section~\ref{sec:diff}, i.e.~cosine distance between average embeddings without significance testing. To set the term frequency threshold to even consider a word for semantic shift detection, we found the optimal threshold that maximised precision@500. This baseline represents the best-case scenario and is, therefore, impossible to replicate in real-world settings as we would not know {\em a priori} which words have undergone semantic shift. The optimal term frequency threshold varied for each simulation run, ranging between 19--25. We refer to this method as {\em Optimal TF}.

\subsection{Results}
\label{sec:simulations}

We evaluated how well FDR adjusted {\em p}-values compare with those from the permutation tests and the baseline. We used precision@$K$ averaged over the 10 simulation runs. Precision@$K = TP(K)/K$, where $K$ is the rank and $TP(K)$ is the count of true positives found in the top-$K$ ranks. We used all values of $K$ from 1--500. 

\subsubsection{Precision@K}

\begin{figure}
    \centering
    \includegraphics[width=\columnwidth]{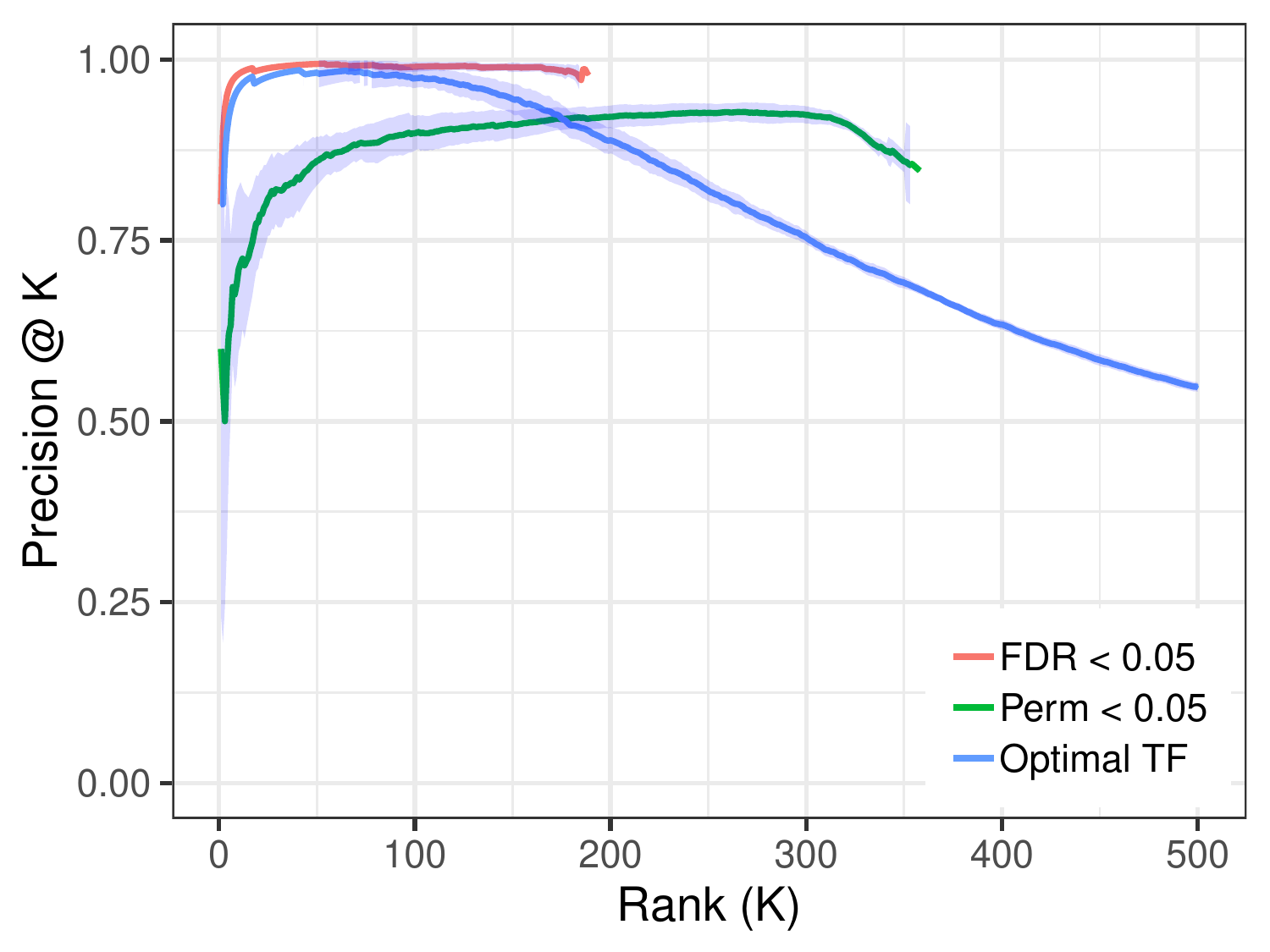}
    \caption{Precision@K averaged over 10 simulation runs. Perm and FDR were filtered using the {\em p}-values from permutation tests and FDR correction, respectively. Optimal TF filters at the optimal term frequency threshold to maximise precision@500.}
    \label{fig:precision}
\end{figure}

Figure~\ref{fig:precision} shows how precision@K varied for the three methods tested. FDR outperformed TF Optimal for all values of $K$ where there was at least $K$ significant results, with the exception of precision@1, where they were equal. FDR output the fewest results of all three methods, ranging from 177 to 189 words across the 10 simulation runs. For precision@177 (the last rank with FDR results in all 10 simulation runs), the lowest precision for FDR was 0.96 compared to 0.86 for TF Optimal, an improvement of 11.6\%. For TF Optimal, performance starts to drop from around rank 100, falling to 0.55 at rank 500. 

We additionally included the {\em p}-values from the permutation tests without FDR correction ({\em Perm} in Figure~\ref{fig:precision}). 
Overall, the permutation tests had the highest precision lower in the ranking, but suffered from more false positives at the top of the ranking compared to the more conservative FDR. 
While FDR and Optimal TF had an average precision@10 close to 1.0, the unadjusted {\em p}-values from the permutation test had an average precision@10 of 0.71. For precision@335 (the last rank with permutation test results in all simulation runs), the lowest precision was 0.86.

\subsubsection{Issues with Term Frequency Thresholds}

\begin{figure}
    \centering
    \includegraphics[width=\columnwidth]{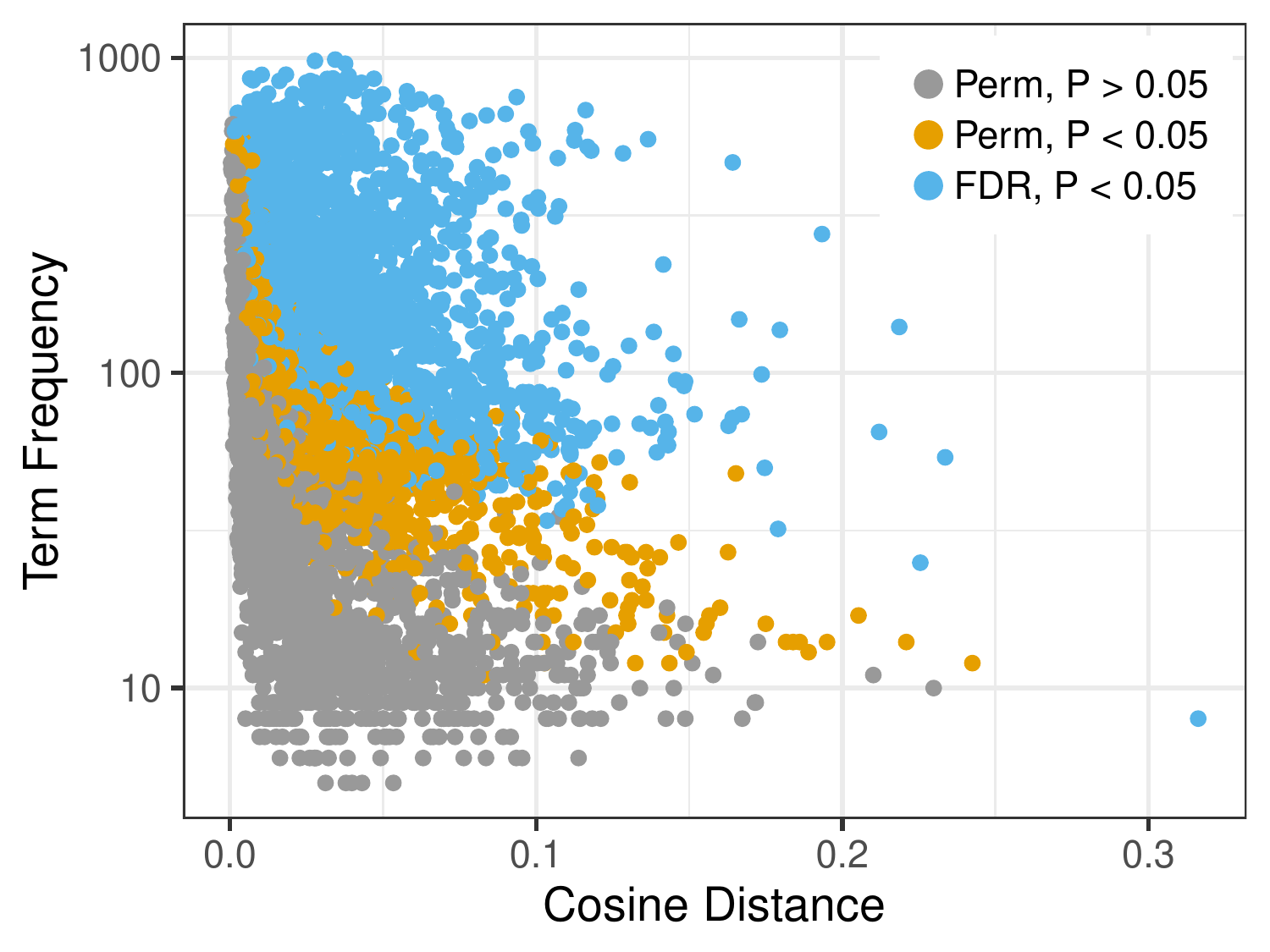}
    \caption{Cosine distance between average BERT embeddings vs term frequency (log scaled). Colors show whether $P < 0.05$ for permutation test and FDR (yellow and blue dots, respectively).}
    \label{fig:significance}
\end{figure}

Figure~\ref{fig:significance} shows the cosine distance for each word with a simulated semantic shift across all 10 simulation runs versus term frequency on a log scale. The colors indicate whether the distance was not statistically significant (gray), significant for the permutation test, but not after FDR correction (yellow) or significant after FDR correction (blue). Significance testing makes a complex trade-off between the effect size (cosine distance), sample size (term frequency) and the number of significance tests performed.
Semantic shift studies usually threshold on term frequency, however, it is clear from Figure~\ref{fig:significance} that thresholding on any combination of term frequency or cosine distance will filter out words where the estimated semantic shift achieves statistical significance, unnecessarily introducing false negatives and potentially omitting important results.

\subsubsection{Factors Influencing Detection}

We hypothesized
that confounding factors, such as the distance between acceptor and donor words in the embedding space, could spuriously influence whether a semantic shift estimate achieves significance. 
To investigate this concern, we created a logistic regression model with a binary response variable of whether the FDR {\em p}-value was $< 0.05$. We used the following explanatory variables from the simulations: gain in term frequency of the acceptor word and two confounding variables 
\begin{enumerate*}[label=(\roman*)]
    \item the final term frequency of the acceptor word and 
    \item the original cosine distance between donor and acceptor words. 
\end{enumerate*}
We standardized all explanatory variables so we could rank them by their relative effect size.
All regression coefficients were highly significant ($P < 2 \times 10^{-16}$). 

The gain in term frequency from donor to acceptor word had the strongest effect on whether semantic change could be detected by FDR ($\beta = 10.7$). The confounding variables, however, had a comparatively modest effect. The term frequency of the acceptor word had a negative effect (-1.52) and the distance between acceptor and donor words had a small positive effect (0.53). 
These findings suggest that significance testing supports semantic change detection and is not strongly biased towards potential confounders.

\section{Empirical Evaluation}

We used five manually annotated data sets (see Table~\ref{tab:data}) to evaluate whether our approach impacts overall performance. We hypothesized that erroneous estimates of semantic change will have a negative impact on the Spearman correlation between cosine distance and manual annotations.


\begin{table*}[]
\small
\centering
\renewcommand*{\arraystretch}{1.2}
\begin{tabular}{lccccccccccc}
\hline
      & \multicolumn{3}{c}{English} & \multicolumn{3}{c}{German} & \multicolumn{3}{c}{Latin} & \multicolumn{2}{c}{Swedish} \\ \cmidrule(lr){2-4} \cmidrule(lr){5-7} \cmidrule(lr){8-10} \cmidrule(lr){11-12}
            & 0.2 & 0.15 & 0.1 & 0.2 & 0.15 & 0.1 & 0.2 & 0.15 & 0.1 & 0.1 & 0.05 \\ \hline
Baseline                & 0.287 & 0.312 & 0.332 & \textbf{0.512} & 0.481 & 0.419              & 0.302             & 0.272 & 0.240 & \textbf{0.134} & 0.141 \\
+ Permutation Tests     & \textbf{0.301}    & 0.330 & \textbf{0.341}    & 0.502 & 0.496 & 0.460             & \textbf{0.304}    & \textbf{0.285} & 0.273 & 0.127 & \textbf{0.162} \\
+ False Discovery Rate  & \textbf{0.301}    & \textbf{0.332} & 0.339    & 0.502 & \textbf{0.498} & \textbf{0.467}     & \textbf{0.304}    & \textbf{0.285} & \textbf{0.286} & 0.127 & \textbf{0.162} \\ \hline
\end{tabular}%
\caption{Spearman correlations of cosine distance vs.~semantic shift for English, German, Latin and Swedish from SemEval-2020. Data sets were randomly sampled without replacement using sampling rates of 0.2, 0.15 and 0.1 (with the exception of Swedish, which was subsampled to 0.1 and 0.05). Each result is the mean over 100 runs.}
\label{tab:semeval}
\end{table*}

\subsection{Experimental Setup}

Each corpus from SemEval-2020 and Liverpool FC was composed of data from two time periods. As in our simulations, we fine-tuned a single BERT model on the entire corpus (i.e.~the concatenation of both time periods) for 5 epochs.
We used the {\tt BERT-base-uncased} pre-trained BERT model for English,  {\tt bert-base-german-dbmdz-uncased} for German, 
 Latin BERT \citep{bamman2020latin} for Latin
and  {\tt af-ai-center/ bert-base-swedish-uncased} for Swedish.
As each corpus provides a list of target words with known semantic shift, we calculated cosine distances, permutation tests and FDR correction for only these words.
We evaluated performance with the Spearman correlation between the estimated semantic shifts and the ground truth. We did this between 
\begin{enumerate*}[label=(\roman*)]
    \item all target words (the baseline),
    \item words where the {\em p}-value from the permutation test was $< 0.05$ and
    \item words where the FDR corrected {\em p}-value was $< 0.05$.
\end{enumerate*}
We note that our chosen baseline \citet{martinc2019leveraging} does not have state-of-the-art performance on SemEval-2020, however, we are only concerned with the relative differences in performance from applying significance testing.

\subsection{SemEval-2020}

Many of the target words in SemEval-2020 have high term frequencies of up to several thousand per time period. 
Given that words with high term frequencies will be highly significant, we created artificially smaller data sets using random subsampling.
For English, German and Latin, we randomly sampled without replacement using sampling rates of 0.2, 0.15 and 0.1. 
For Swedish, we used sampling rates of 0.1 and 0.05. 
These sampling rates were chosen to cover the range of almost all words achieving significance to only $\mathord{\sim}30$ achieving significance (below which, the Spearman correlation was often not statistically significant). Swedish has only 31 words in the target set and the highest term frequencies, necessitating lower sampling rates. 
We created 100 randomly subsampled data sets per sampling rate for each language. 
Model fine-tuning took approximately $1, 24, 2.5$ and $36$ hours for English, German, Latin and Swedish, respectively, using the same computing resources as previously. {\em P}-value calculation varied per language and sampling rate from 25 minutes to 2 hours.

\subsubsection{Results}

Table~\ref{tab:semeval} shows the average Spearman correlation for each data set in SemEval-2020 for each sampling rate tested. 
Despite the small size of the target sets, significance testing generally had a positive impact on Spearman correlation.
FDR correction had the highest or joint highest correlation in 8/11 experiments.
The baseline outperformed both permutation tests and FDR correction in two experiments, however, the difference was at most only 0.01.
At the lowest sampling rates, the margin by which our approach outperformed the baseline varied widely: with a sampling rate of 0.1 there was an improvement of 11.5\% and 19.2\%, for German and Latin, respectively. However, for English the improvement was only 2.5\%. 


In these experiments, performing FDR correction, compared to permutation tests, made a minimal difference to correlation because there were only 31-48 significance tests compared to $\mathord{\sim}31,000$ simultaneous tests performed in Section~\ref{sec:syneval}. 

\subsection{Liverpool FC Subreddit}

Compared to SemEval-2020, words in the Liverpool FC corpus tend to have lower term frequencies. Additionally, the Liverpool FC corpus had a higher proportion of words with a ground truth semantic shift of zero (40/97). The Liverpool FC corpus better represents the kind of observational studies we  focus on with our approach, where data is sparse and therefore noisier.
Model fine-tuning and {\em p}-value calculation took $< 3$ hours in total.

\subsubsection{Results}

\begin{figure}
    \centering
    \includegraphics[width=\columnwidth]{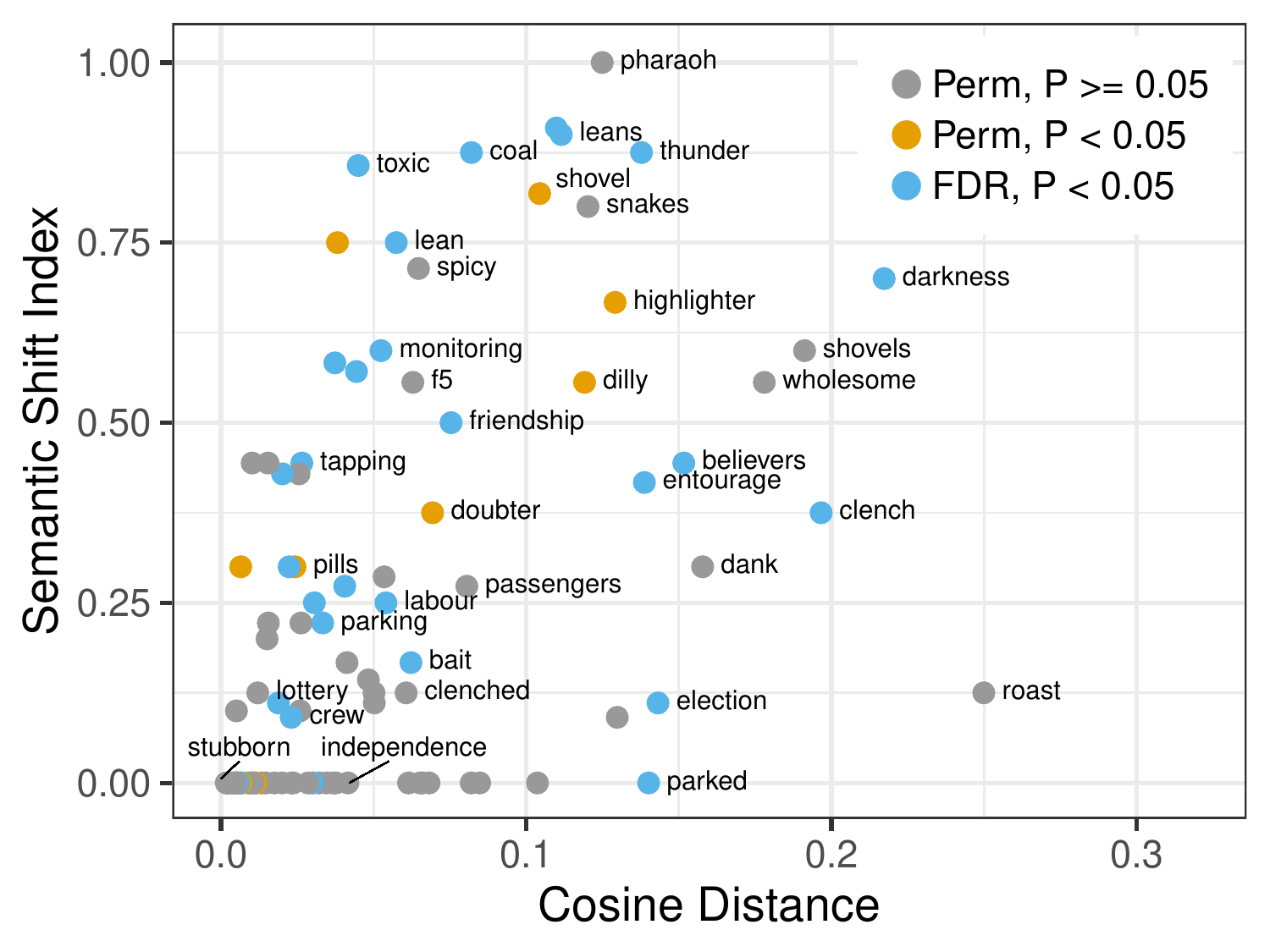}
    \caption{Cosine distance vs.~semantic shift index for Liverpool FC data set. Colors show whether $P < 0.05$ for permutation test and FDR (yellow and blue dots).}
    \label{fig:lfcwords}
\end{figure}

Figure~\ref{fig:lfcwords} shows a scatter plot of the cosine distance versus semantic shift index for the 97 words in the target set. 
Each word is colored by {\em p}-value: gray were not statistically significant (57/97), yellow were significant for the permutation test, but not after FDR correction (10/97) and blue were significant after FDR correction (30/97). 
For a majority of words, there is insufficient evidence to reject the null hypothesis that the semantic change is zero.

Some words failed to achieve significance due to low term frequency: {\em dank} and {\em roast}, for example, only occur once in the 2011-2013 corpus. 
Other words had a high term frequency, but a low estimated shift -- requiring substantial evidence to be considered significantly different from zero. 
Of the 40 target words with a ground truth semantic shift of zero, FDR correction found that 35 were not significantly different from zero (compared to 32 using raw {\em p}-values from the permutation test). 

The number of words filtered out by significance testing is so great that it negatively impacts correlation. Table~\ref{tab:lfc} shows the Spearman correlations. The method with the highest correlation was using permutation tests without FDR correction, outperforming the baseline by 4.7\%. 
The loss in correlation in FDR correction is caused by three outliers ({\em clench, election} and {\em parked}). The negative influence of these words would be lessened if more words had achieved statistical significance, as was the case with using just permutation tests.

Significance testing highlights how we can be misled by randomness. 
\citet{del2018short} stated that false positives with a semantic shift index of zero may be caused by a referential effect, i.e.~words that refer to different people or events in different time periods, such as 
{\em independence} which referred to events in Catalonia (Figure~\ref{fig:lfcwords}, bottom left). 
However, a simpler explanation is that there is insufficient evidence to support the cosine distance being significantly different from zero. 
Similarly, the estimates for {\em pharaoh} and {\em shovel} are not too low because they are being used metaphorically \citep{del2018short}, but because the estimated semantic shift is not statistically significant.

\section{Conclusions}


We presented an approach to identify statistically significant semantic change using contextual word embeddings, permutation tests and false discovery rate. Our work was motivated by the fact that while there are many methods for semantic change detection, they only estimate the magnitude of a given shift and ignore the uncertainty in their estimates. 
As a result, existing semantic change detection methods are problematic to apply if 
\begin{enumerate*}[label=(\roman*)]
    \item the data set is of limited size or
    \item you need to estimate the semantic shift for words with low term frequencies.
\end{enumerate*}

In simulation, we demonstrated that using a combination of permutation tests and false discovery rate allows us to scale semantic shift estimation to every word in the vocabulary while avoiding false positives and achieving high precision (Figure~\ref{fig:precision}).
In our analysis of the SemEval-2020 data sets, we showed that significance testing has a generally positive impact on Spearman correlation (Table~\ref{tab:semeval}). In these examples, the false discovery rate procedure made less of a difference due to the limited number of significance tests performed.
In the Liverpool FC corpus, less than a third of target words achieved significance with FDR correction, causing outliers to have a greater influence on correlation than they did on permutation tests (Table~\ref{tab:lfc}).
However, we were able to highlight how conclusions from \citet{del2018short} could be more easily explained by semantic shift estimates not being statistically significant due to insufficient evidence.


\begin{table}[]
\small
\centering
\renewcommand*{\arraystretch}{1.2}
\begin{tabular}{lc}
\hline
         & \multicolumn{1}{l}{\begin{tabular}[c]{@{}l@{}}Liverpool FC\end{tabular}} \\ \hline
Baseline & 0.536 \\
+ Permutation Tests & \textbf{0.561} \\
+ False Discovery Rate & 0.478 \\ \hline
\end{tabular}%
\caption{Spearman correlations of cosine distance vs.~semantic shift for the Liverpool FC corpus.}
\label{tab:lfc}
\end{table}

Our work has several limitations. First, the need to recalculate semantic shifts for each permutation increases computational requirements substantially. While this did not prevent us from calculating {\em p}-values for the whole vocabulary in our simulations, it will be challenging to scale to more complex methods that are based on, for example, clustering.
Second, false discovery rate assumes that simultaneous significance tests are independent, which is not the case for word usage. A better approach would account for correlations in usage between words to calibrate the significance threshold more appropriately.

In this paper, we used a semantic shift method which did not exploit the full potential of contextual word embeddings, i.e.~to disambiguate between word senses. 
%
In future work, we are going to investigate whether more scalable techniques (e.g.~\citet{dwass1957modified}) could be applied at the word sense level, where even large data sets could potentially start to suffer from the small data problems we focused on in this article.

\section*{Acknowledgements}
The authors were supported by a grant from Business Finland grant number 3283/31/2019.
The authors wish to acknowledge CSC – IT Center for Science, Finland, for computational resources.


\bibliography{anthology,custom}
\bibliographystyle{acl_natbib}

\end{document}